%% file: main.tex
\documentclass[runningheads]{llncs}
\usepackage{graphicx}
\usepackage{booktabs}
\usepackage{subfigure}
\usepackage{tikz}
\usepackage{array}
\usepackage[ruled,linesnumbered]{algorithm2e}
\usepackage{algpseudocode}
\usepackage{comment}
\usepackage{amsmath,amssymb} 
\usepackage{color}
\usepackage{multirow}
\usepackage{amssymb}
\usepackage{pifont}
\newcommand{\cmark}{\ding{51}}%
\newcommand{\xmark}{\ding{55}}%
\usepackage[accsupp]{axessibility}  

\usepackage[breaklinks,colorlinks]{hyperref}

\makeatletter
  \newcommand\figcaption{\def\@captype{figure}\caption}
  \newcommand\tabcaption{\def\@captype{table}\caption}
\makeatother

\input{defs.tex}

\begin{document}

\pagestyle{headings}
\mainmatter
\def\ECCVSubNumber{444}  

\title{Fusing Local Similarities for Retrieval-based 3D Orientation Estimation of Unseen Objects}

\titlerunning{Unseen Object Orientation Estimation}

\author{Chen Zhao$^{1}$\and
Yinlin Hu$^{1,2}$ \and
Mathieu Salzmann$^{1,2}$}

\authorrunning{Zhao et al.}

\institute{$^{1}$EPFL-CVLab, $^{2}$ClearSpace SA \\
\email{\{chen.zhao, yinlin.hu, mathieu.salzmann\}@epfl.ch}}

\maketitle

\begin{abstract}
In this paper, we tackle the task of estimating the 3D orientation of previously-unseen objects from monocular images. This task contrasts with the one considered by most existing deep learning methods which typically assume that the testing objects have been observed during training. To handle the unseen objects, we follow a retrieval-based strategy and prevent the network from learning object-specific features by computing multi-scale local similarities between the query image and synthetically-generated reference images. We then introduce an adaptive fusion module that robustly aggregates the local similarities into a global similarity score of pairwise images. Furthermore, we speed up the retrieval process by developing a fast retrieval strategy. Our experiments on the LineMOD, LineMOD-Occluded, and T-LESS datasets show that our method yields a significantly better generalization to unseen objects than previous works. Our code and pre-trained models are available at \url{https://sailor-z.github.io/projects/Unseen_Object_Pose.html}.
\end{abstract}

\section{Introduction}
\label{sec:intro}
Estimating the 3D orientation of objects from an image is pivotal to many computer vision and robotics tasks, such as robotic manipulation~\cite{collet2011moped,zhu2014single,tremblay2018deep}, augmented reality, and autonomous driving~\cite{geiger2012we,chen2017multi,xu2018pointfusion,marchand2015pose}. Motivated by the tremendous success of deep learning, much effort~\cite{xiang2017posecnn,peng2019pvnet,wang2019densefusion} has been dedicated to developing deep networks able to recognize the objects depicted in the input image and estimate their 3D orientation. To achieve this, most learning-based methods assume that the training data and testing data contain exactly the same objects~\cite{hu2020single,wang2021gdr} or similar objects from the same category~\cite{wang2019normalized,manuelli2019kpam}. 
However, this assumption is often violated in real-world applications, such as robotic manipulation, where one would typically like the robotic arm to be able to handle previously-unseen objects without having to re-train the network for them. 

\begin{figure}[t]
	\centering
	\includegraphics[width=0.8\linewidth]{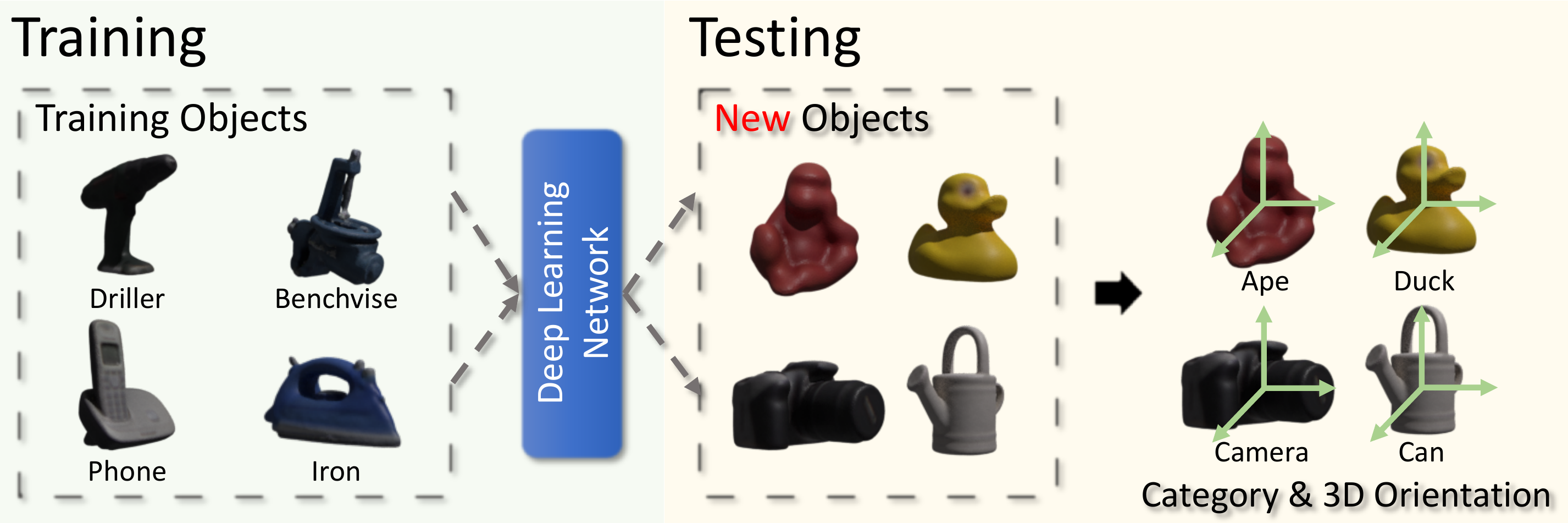}
	\caption{\textbf{3D orientation estimation for unseen objects.} The network is trained on a limited number of objects and tested on unseen (new) objects that fundamentally differ from the training ones in shape and appearance. The goal is to predict both the category and the 3D orientation of these unseen objects.}
	\label{fig:intro}
\end{figure}

In this paper, as illustrated in Fig.~\ref{fig:intro}, we tackle the task of 3D orientation estimation for \emph{previously-unseen} objects. Specifically, we develop a deep network that can be trained on a limited number of objects, and yet remains effective when tested on novel objects that drastically differ from the training ones in terms of both appearance and shape. To handle such previously-unseen objects, we cast the task of 3D orientation estimation as an image retrieval problem. We first create a database of synthetic images depicting objects in different orientations. Then, given a real query image of an object, we search for the most similar reference image in the database, which thus indicates both the category and 3D orientation of this object. 

Intuitively, image retrieval methods~\cite{wohlhart2015learning} offer a promising potential for generalization, because they learn the relative similarity of pairwise images, which can be determined without being aware of the object category. However, most previous works~\cite{wohlhart2015learning,sundermeyer2018implicit,arandjelovic2016netvlad,vaze2022generalized,xiao2021posecontrast} that follow this approach exploit a global image representation to measure image similarity, ignoring the risk that a global descriptor may integrate high-level semantic information coupled with the object category, which could affect the generalization ability to unseen objects. To address this problem, our approach relies on the similarities of local patterns, which are independent to the object category and then facilitate the generalization to new objects. Specifically, we follow a multi-scale strategy
and extract feature maps of different sizes from the input image. To facilitate the image comparison process, we then introduce a similarity fusion module, adaptively aggregating multiple local similarity scores into a single one that represents the similarity between two images. To further account for the computational complexity of the resulting multi-scale local comparisons, we design a fast retrieval strategy.

We conduct experiments on three public datasets, LineMOD~\cite{hinterstoisser2012model}, LineMOD-Occluded (LineMOD-O)~\cite{brachmann2014learning}, and T-LESS~\cite{hodan2017t}, comparing our method with both hand-crafted~\cite{dalal2005histograms} and deep learning~\cite{wohlhart2015learning,arandjelovic2016netvlad,xiao2019pose,sundermeyer2020multi} approaches. Our empirical results evidence the superior generalization ability of our method to previously-unseen objects. Furthermore, we perform ablation studies to shed more light on the effectiveness of each component in our method. Our contributions can be summarized as follows:
\begin{itemize}
    \item We estimate the 3D orientation of previously-unseen objects by introducing an image retrieval framework based on multi-scale local similarities.
    
    \item We develop a similarity fusion module, robustly predicting an image similarity score from multi-scale pairwise feature maps.
    
    \item We design a fast retrieval strategy that achieves a good trade-off between the 3D orientation estimation accuracy and efficiency.
\end{itemize}

\section{Related Work}
\label{sec:related}
\noindent\textbf{Object Pose Estimation.} In recent years, deep learning has been dominating the field of object pose estimation. For instance, PoseCNN~\cite{xiang2017posecnn} 
relies on two branches to directly predict the object orientation as a quaternion, and the 2D location of the object center, respectively. PVNet~\cite{peng2019pvnet} estimates the 2D projections of 3D points using a voting network. The object pose is then recovered by using a PnP algorithm~\cite{fischler1981random} over the predict 2D-3D correspondences. DenseFusion~\cite{wang2019densefusion} fuses 2D and 3D features extracted from RGB-D data, from which it predicts the object pose. GDR-Net~\cite{wang2021gdr} predicts a dense correspondence map, acting as input to a Patch-PnP module that recovers the object pose. These deep learning methods have achieved outstanding pose estimation accuracy when the training and testing data contain the same object instances~\cite{du2021vision}. However, the patterns they learn from the input images are instance specific, and these methods cannot generalize to unseen objects~\cite{park2020latentfusion}.
\\

\noindent\textbf{Category-Level Object Pose Estimation.} Some methods nonetheless loosen the constraint of observing the same object instances at training and testing time by performing category-level object pose estimation~\cite{wang2019normalized,chen2021fs,wang20206,li2020category}. These methods assume that the training data contain instances belonging to a set of categories, and new instances from these categories are observed during testing. In this context, a normalized object coordinate space (NOCS) is typically used~\cite{wang2019normalized,li2020category}, providing a canonical representation shared by different instances within the same category. The object pose is obtained by combining the NOCS maps, instance masks, and depth values. These methods rely on the intuition that the shapes of different instances in the same category are similar, and then the patterns learned from the training data can generalize to new instances in the testing phase. As such, these methods still struggle in the presence of testing objects from entirely new categories. Furthermore, all of these techniques require \emph{depth} information as input. By contrast, our method relies only on RGB images, and yet can handle unseen objects from new categories at testing time.
\\

\noindent\textbf{Unseen Object Pose Estimation.} A few attempts at predicting the pose of unseen objects have been made in the literature. In particular, LatentFusion~\cite{park2020latentfusion} introduces a latent 3D representation and optimizes an object's pose by differentiable rendering. DeepIM~\cite{li2018deepim} presents an iterative framework, using a matching network to optimize an initial object pose. Both of these methods require an \emph{initial} pose estimate, which is typically hard to obtain for unseen objects. Furthermore, the pose estimation step in LatentFusion leverages \emph{depth} information. Since estimating the full 6D pose of an \emph{unseen} object from a single RGB image is highly challenging, several works suggest simplifying this problem by focusing on estimating the 3D object orientation~\cite{wohlhart2015learning,balntas2017pose,sundermeyer2018implicit,xiao2019pose,xiao2021posecontrast}. These methods utilize the 3D object model to generate multi-view references, which are then combined with the real image to either perform template matching~\cite{wohlhart2015learning,balntas2017pose,sundermeyer2018implicit,sundermeyer2020multi,xiao2021posecontrast} or directly regress the 3D orientation~\cite{xiao2019pose}. However, they propose to learn a global representation from an image, in which the high-level semantic information is correlated to the object and then limits the generalization to unseen objects. 
In this paper, we also focus on 3D orientation estimation of unseen objects, but handle this problem via a multi-scale local similarity learning network.

\section{Method}
\label{sec:method}
\begin{figure}[!t]
	\centering
	\subfigure[Previous work]
	{ \includegraphics[width=0.33\linewidth]{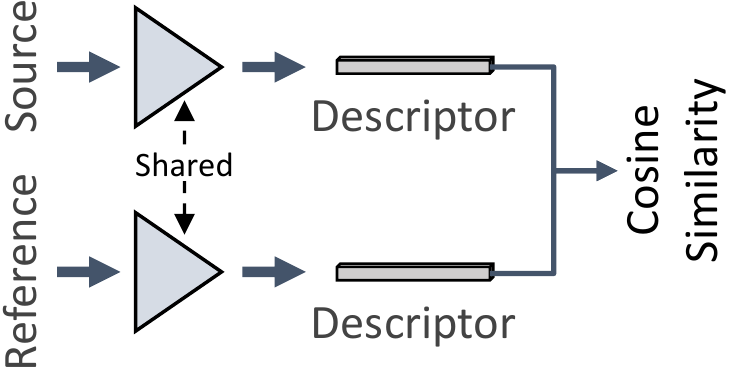}\label{fig:moti_a}
	}
	\hspace{3em}
	\subfigure[Our method]
	{\includegraphics[width=0.4\linewidth]{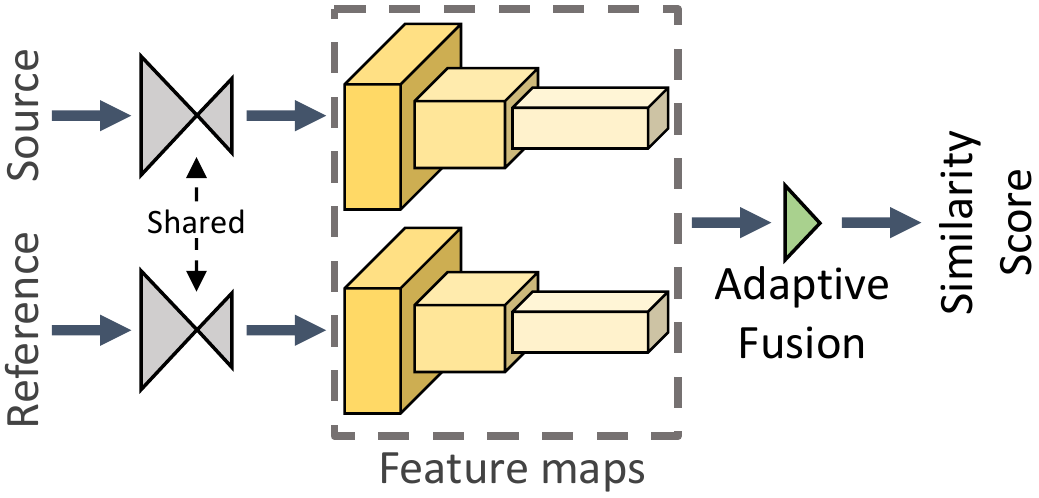}\label{fig:moti_b}
	}
	\caption{\textbf{Difference between previous works and our method.} (a) Existing works convert images into global descriptors that are used to compute a similarity score for retrieval. (b) Our method compares images using local similarities between the corresponding elements in feature maps, and adaptively fuses these local similarities into a single one that indicates the similarity of two images.}
	\label{fig:moti}
\end{figure}

\subsection{Problem Formulation}
Let us assume to be given a set of training objects $\mathcal{O}_{train}$ belonging to different categories $\mathcal{C}_{train}$.\footnote{In our scenario, and in contrast to category-level pose estimation, each object instance corresponds to its own category.} As depicted by Fig.~\ref{fig:intro}, we aim to train a model that can predict the 3D orientation of new objects $\mathcal{O}_{test}$, $\mathcal{O}_{test}\cap\mathcal{O}_{train}=\emptyset$, from entirely new categories $\mathcal{C}_{test}$, $\mathcal{C}_{test}\cap\mathcal{C}_{train}=\emptyset$. Specifically, given an RGB image $\mathbf{I}_{src}$ containing an object $O_{src} \in \mathcal{O}_{test}$, our goal is to both recognize the object category $C_{src}$ and estimate the object's 3D orientation, expressed as a rotation matrix $\mathbf{R}_{src}\in\mathbb{R}^{3\times3}$. We tackle this dual problem as an image retrieval task. For each $O_{src}^{i}\in(\mathcal{O}_{train}\cup\mathcal{O}_{test})$, we generate references $\mathcal{I}_{ref}^{i}$ with different 3D orientations by rendering the corresponding 3D model $\mathbf{M}_i$. We then seek to pick $\hat{\mathbf{I}}_{ref} \in \{\mathcal{I}_{ref}^{1}\cup\mathcal{I}_{ref}^{2}\cdots\cup\mathcal{I}_{ref}^{N}\}$ that is the most similar to $\mathbf{I}_{src}$. The category label $C_{src}$ and 3D orientation $\mathbf{R}_{src}$ of $O_{src}$ are then taken as those of the corresponding $\hat{O}_{ref}$.

\subsection{Motivation}
As illustrated in Fig.~\ref{fig:moti_a}, the existing retrieval-based 3D orientation estimation methods~\cite{wohlhart2015learning,sundermeyer2018implicit,xiao2021posecontrast} convert an image into a global descriptor. Retrieval is then performed by computing the similarity between pairs of descriptors. As a consequence, the deep network that extracts the global descriptor typically learns to encode object-specific semantic information in the descriptor, which results in a limited generalization ability to unseen objects. By contrast, we propose to compare images via local patch descriptors, in which it is harder to encode high-level semantic information thus encouraging the network to focus on local geometric attributes. As shown in Fig.~\ref{fig:moti_b}, we estimate local similarities between the corresponding elements in source and reference feature maps. Furthermore, to enforce robustness to noise, such as background, we introduce an adaptive fusion module capable of robustly predicting an image similarity score from the local ones.

\subsection{Multi-scale Patch-level Image Comparison}
\label{sec:comparison}
\begin{figure}[!t]
	\centering
	\includegraphics[width=1.0\linewidth]{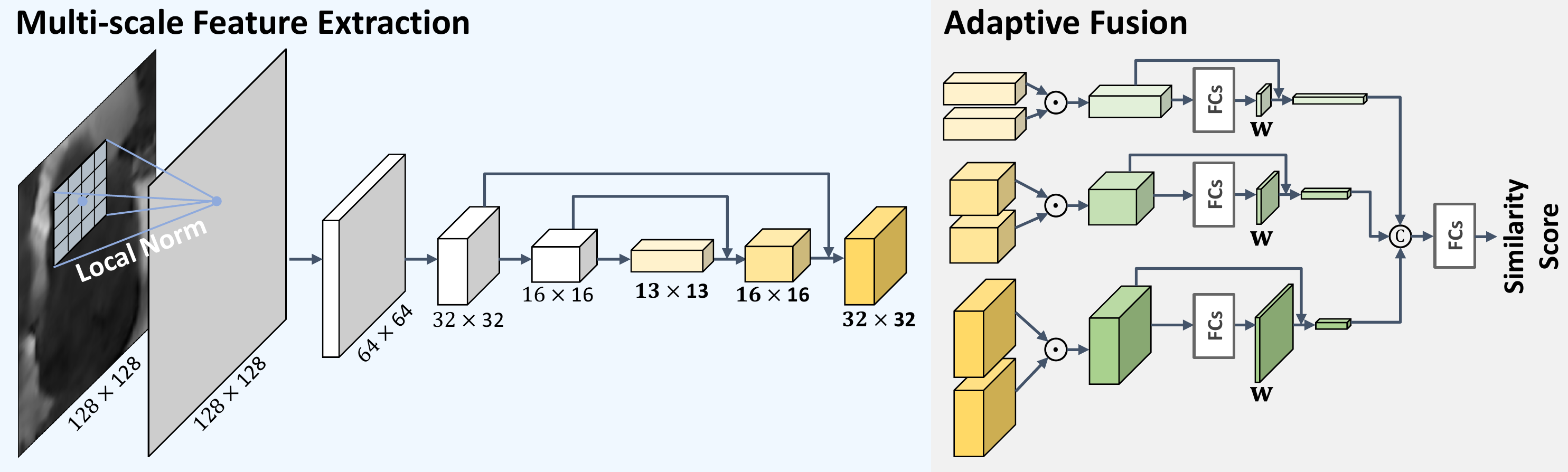}
	\caption{\textbf{Network architecture.} We extract multi-scale features from a locally-normalized image. We then compute local similarities at each scale between the features of the source image and those of a reference one, and adaptively fuse them into a global similarity score. 
	}
	\label{fig:network}
	\vspace{-15pt}
\end{figure}
As the source image is real but the reference ones are synthetic with discretely sampled 3D orientation, the appearance and shape variations are inevitable even for the most similar reference. Moreover, the background included in the source image, but absent from the reference ones, typically interferes with our patch-level comparisons. In practice, we have observed that small patches could be too sensitive to appearance and shape variations, while large patches tend to be affected by the background. Finding a single effective patch size balancing robustness to the domain gap and to the background therefore is challenging. 

To address this issue, we introduce a multi-scale feature extraction module. As shown in Fig.~\ref{fig:network}, our network takes a grey-scale image $\mathbf{I}\in\mathbb{R}^{128\times128}$ as input, which shows better robustness than color images in practice. 
Subsequently, we employ a series of ResNet layers~\cite{he2016deep}, estimating a down-sampled feature map $\mathbf{F}\in\mathbb{R}^{13\times13\times C}$. We compute multi-scale feature representations by progressively up-sampling $\mathbf{F}$ using deconvolution layers~\cite{long2015fully} and bilinear interpolation. We also utilize skip connections~\cite{ronneberger2015u} to better preserve the geometric information. The elements in the generated multi-scale feature maps then encode patches of different sizes in $\mathbf{I}$, which enables multi-scale patch-level image comparison.

To perform image retrieval, one nonetheless needs to compute a single similarity score for a pair of images. To this end, we compare the pairwise multi-scale feature maps and fuse the resulting local similarities into a single score expressed as
\begin{align}
\label{eq:eq3}
s=f(g(\mathbf{F}_{src}^{1}, \mathbf{F}_{ref}^{1}), g(\mathbf{F}_{src}^{2}, \mathbf{F}_{ref}^{2}), \cdots, g(\mathbf{F}_{src}^{S}, \mathbf{F}_{ref}^{S})),
\end{align}
where $\mathbf{F}_{src}$ and $\mathbf{F}_{ref}$ represent the feature maps of $\mathbf{I}_{src}$ and $\mathbf{I}_{ref}$, respectively, and $S$ denotes the number of scales. A straightforward solution to estimate $s$ is to compute the per-element cosine similarity for all pairs $(\mathbf{F}_{src}^{i},\mathbf{F}_{ref}^{i})$, with $i\in\{1, 2,\cdots, S\}$, and average the resulting local similarities. However, this strategy would not be robust to outlier patches, such as those dominated by background content. Therefore, we introduce an adaptive fusion strategy illustrated in the right part of Fig.~\ref{fig:network}. Following the same formalism as above, it computes an image similarity score as
\begin{align}
\label{eq:eq4}
s=f(\text{cat}\left[g({\mathbf{F}}^{*}_{i}\odot\mathbf{w}_{i})\right], \psi),i\in\{1, 2,\cdots,S\},
\end{align}
where $cat$ indicates the concatenation process, $g:\mathbb{R}^{H\times W \times C} \rightarrow \mathbb{R}^{C}$ denotes the summation over the spatial dimensions, $\mathbf{F}^{*}_{i}$ represents the local similarities obtained by computing the cosine similarities between the corresponding elements in $\mathbf{F}_{src}^{i}$ and $\mathbf{F}_{ref}^{i}$, $\odot$ indicates the Hadamard product, and $\psi$ represents the learnable parameters of the fully connected layers (FCs) $f(\cdot)$. 
The weights ${\bf w}_i$ encode a confidence map over ${\mathbf{F}}^{*}_{i}$ to account for outliers, and are computed as
\begin{align}
\label{eq:eq6}
\mathbf{w}_{i}=\frac{\text{exp}(h({\mathbf{F}}^{*}_{i}, \mathbf{\omega}))\odot\text{sigmoid}(q({\mathbf{F}}^{*}_{i}, \mathbf{\theta}))}{\sum\text{exp}(h({\mathbf{F}}^{*}_{i}, \mathbf{\omega}))\odot\text{sigmoid}(q({\mathbf{F}}^{*}_{i}, \mathbf{\theta}))},
\end{align}
where $\omega$ and $\theta$ are learnable parameters of the convolutional layers $h(\cdot)$ and $q(\cdot)$, respectively. 
This formulation accounts for both the individual confidence of each element in ${\mathbf{F}}^{*}_{i}$ via the sigmoid function, and the relative confidence w.r.t. all elements jointly via the softmax-liked function. As such, it models both the local and global context of ${\mathbf{F}}^{*}_{i}$, 
aiming to decrease the confidence of the outliers while increases that of the inliers. Our experimental results in Section~\ref{sec:abl} show that our adaptive fusion yields better results than the straightforward averaging process described above, even when trained in an unsupervised manner. 

To further reduce the effects of object-related patterns in local regions and synthetic-to-real domain gap, we pre-process $\mathbf{I}$ via a local normalization. Each pixel $\overline{p}_{ij}$ in the normalized image is computed as
\begin{align}
\label{eq:eq1}
\overline{p}_{ij} = \frac{p_{ij}-\mu}{\sigma}, 
\end{align}
where $p_{ij}$ is the corresponding pixel in $\mathbf{I}$, and
\begin{align}
\label{eq:eq2}
\mu = \frac{1}{r^{2}}\sum_{i^{'},j^{'}}p_{i^{'}j^{'}},\;\;\sigma=\sqrt{\frac{1}{r^{2}}\sum_{i^{'},j^{'}}{(p_{i^{'}j^{'}}-\mu)}^{2}},
\end{align}
with $i^{'}\in\left[i\pm r/2\right]$, $j^{'}\in\left[j\pm r/2\right]$, and $r$ denoting the window size. 
\subsection{Fast Retrieval}
\label{sec:fast}
Although the proposed patch-level image comparison integrates more local geometric information than the image-level methods~\cite{wohlhart2015learning,sundermeyer2018implicit}, and as will be shown by our experiments thus yields better generalization to unseen objects, it suffers from a high retrieval time. Indeed, a na\"ive image retrieval strategy compares $\mathbf{I}_{src}$ with every reference in the database. Given $N$ objects with $R$ reference images each, the cost of $O(NR)$ quickly becomes unaffordable as $N$ and $R$ increase. This could be remedied by parallel computing, but at the cost of increasing memory consumption. Here, we therefore introduce a fast retrieval method that balances effectiveness and efficiency.

\begin{figure}[!t]
	\centering
	\includegraphics[width=0.8\linewidth]{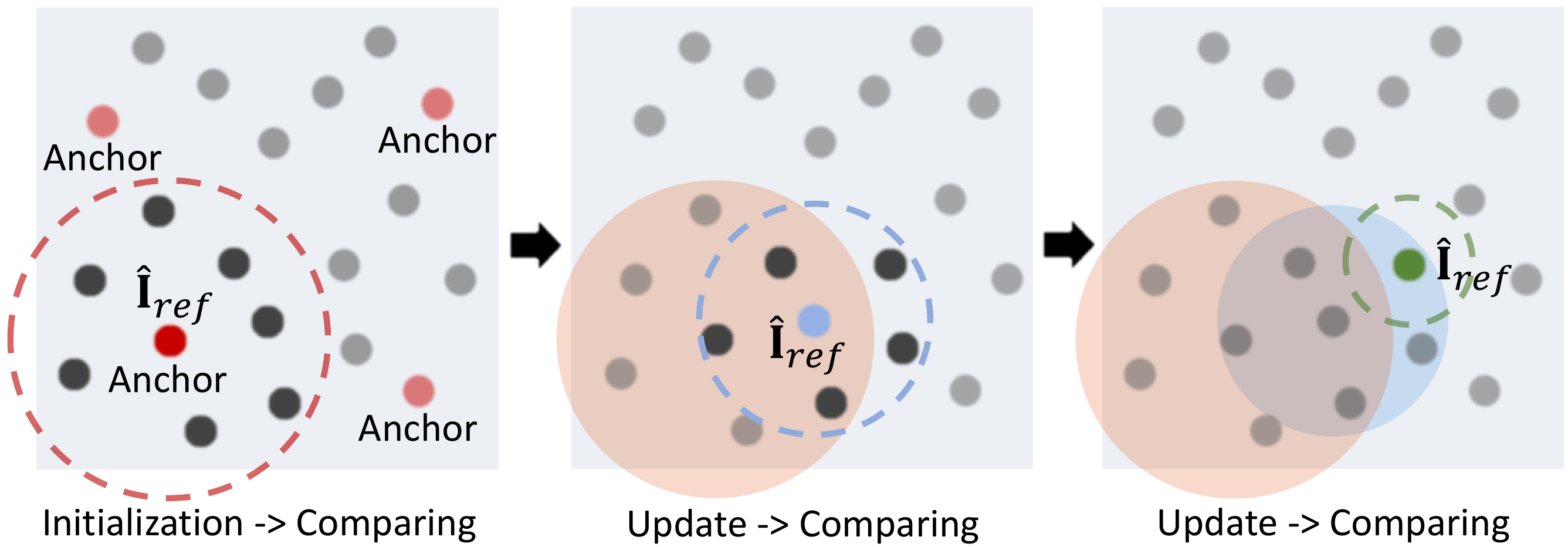}
	\caption{\textbf{Fast retrieval.} The location of the source image in the reference database is initialized by comparing the source image with a set of anchors. $\hat{\mathbf{I}}_{ref}$ is dynamically updated based on the similarity scores of the references within a local region around the current $\hat{\mathbf{I}}_{ref}$ estimate.}
	\label{fig:retrieval}
\end{figure}

As illustrated in Fig.~\ref{fig:retrieval}, instead of comparing $\mathbf{I}_{src}$ with all the references one-by-one, we first roughly locate $\mathbf{I}_{src}$ in the reference database and then iteratively refine this initial location. Our method is summarized in Algorithm~\ref{algo:fast}. We omit some subscripts for convenience. Specifically, for each object, we first sample $k_{ac}$ anchors from $\mathcal{I}_{ref}$ using farthest point sampling (FPS), which leads to a good coverage~\cite{qi2017pointnet++} of $\mathcal{I}_{ref}$. This is done using the geodesic distance of the corresponding 3D rotation matrices as a metric in FPS. The anchor with the largest score $s$ computed from Eq.~\ref{eq:eq4} is taken as the initial point $\hat{\mathbf{I}}_{ref}$. Subsequently, we perform retrieval using the method described in Section~\ref{sec:comparison} within a local region centered at the current $\hat{\mathbf{I}}_{ref}$, and update $\hat{\mathbf{I}}_{ref}$ based on the similarity scores. Such updates are performed until convergence. 

The straightforward application of this strategy would be prone to local optima. Intuitively, this can be addressed by increasing the size of the local region, but this would come at a higher computational cost. Therefore, we determine a search space and further make use of FPS to select anchors within the space, because FPS covers a larger region than KNN with the same number of samples. At each iteration, we decrease the radius of the search space to further improve efficiency.

\begin{algorithm}[t]
	\caption{Fast Retrieval}
	\label{algo:fast}
	\LinesNumbered
	\KwIn{$\mathbf{I}_{src}$, $\mathcal{I}_{ref}$, $\mathcal{R}_{ref}=\{\textbf{R}_1,\textbf{R}_2,\cdots,\textbf{R}_R\}, k_{ac}, R$}
	\KwOut{$\hat{\mathbf{I}}_{ref}$, $\mathbf{R}_{src}$}
	Sample $k_{ac}$ anchors from $\mathcal{I}_{ref}$ using FPS; \\
	Estimate similarities using Eq.~\ref{eq:eq4}; \\
	Initialize $\hat{\mathbf{I}}_{ref}$ as the most similar anchor; \\
	$j=1$; \\
	\Repeat{$\hat{\mathbf{I}}_{ref}$ converges}{
	Define a search space around $\hat{\mathbf{I}}_{ref}$ with a radius of $\lfloor{R/2^{j}}\rfloor$; \\
	
	Compute anchors using FPS; \\
	Estimate similarities using Eq.~\ref{eq:eq4}; \\
	Update $\hat{\mathbf{I}}_{ref}$; \\
	$j$++;
	}
	Determine $\mathbf{R}_{src}$ as $\hat{\mathbf{R}}_{ref}\in\mathcal{R}_{ref}$.
\end{algorithm}

\subsection{Training and Testing}
In the training stage, we follow the infoNCE contrastive learning formalism~\cite{chen2020simple}. We associate each training sample (source image) in a mini-batch with its closest reference to form a positive pair. To better cover the entire training set, we also group each positive pair with a random sample, leading to a triplet. Note that in the standard infoNCE loss~\cite{chen2020simple} for contrastive learning, all samples except for the most similar one are treated as negative. In our context, all reference images would be penalized equally, except for the one from the same category and with the closest 3D orientation. Their 3D orientation difference is thus not differentiated.
To better account for the continuous nature of the 3D orientation difference and avoid over-penalizing some references, we introduce a weighted-infoNCE loss
\begin{align}
\label{eq:eq7}
\ell_{ij} = -\text{log}\frac{\text{exp}(s_{ij} / \tau)\cdot w_{ij}}{\sum_{k=1}^{3B}\text{exp}(s_{ik} / \tau)\cdot w_{ik}},
\end{align}
where $s_{ij}$ is the similarity score of the positive pair $(\mathbf{I}_i,\mathbf{I}_j)$, $\tau=0.1$ denotes a temperature parameter~\cite{chen2020simple}, $B$ is the size of a mini-batch, and $w_{ij}$ (and similarly $w_{ik}$) represents a weight computed as
\begin{equation}
\label{eq:eq8}
w_{ij}=\left\{
\begin{array}{lcl}
\text{arccos}(\frac{\text{tr}(\mathbf{R}_{i}^{\text{T}}\mathbf{R}_{j})-1}{2})/\pi &  \ \ \ \ & \text{if} \ C_{j}=C_{i} \\
1 & \ \ \ \ & \text{else}
\end{array} \right..
\end{equation}

In the testing phase, we store the features of the reference images and sample the initial anchors offline; we recognize the object categories and predict the 3D orientation online. More specifically, we first compare $\mathbf{I}_{src}$ with $k_{ac}$ anchors for each $\mathcal{I}_{ref}^{i}$, and take $C_{src}$ to be the category of the anchor with the largest similarity score. This process reduces the complexity of object category recognition from $O(NR)$ to $O(Nk_{ac})$. Subsequently, we restrict the search for $\hat{\mathbf{I}}_{ref}$ in our fast retrieval process to the references depicting the recognized object. $\mathbf{R}_{src}$ is finally taken as the corresponding rotation matrix $\hat{\mathbf{R}}_{ref}$.
\section{Experiments}
\label{sec:exper}
\subsection{Implementation Details}
In our experiments, we set the window size for local normalization and the number of anchors for fast retrieval to $r=32$ and $k_{ac}=1024$, respectively. We train our network for $200$ epochs using the Adam~\cite{kingma2014adam} optimizer with a batch size of $16$ and a learning rate of $10^{-4}$, which is divided by 10 after $50$ and $150$ epochs. Training takes around $20$ hours on a single NVIDIA Tesla V100.

\subsection{Experimental Setup}
Following the standard setting~\cite{wohlhart2015learning,sundermeyer2018implicit,xiao2019pose,sundermeyer2020multi}, we assume to have access to 3D object models, which provide us with canonical frames, without which the object orientation in the camera frame would be ill-defined.
Note that the 3D object models are also used to generate the reference images.
We generate $R=10,000$ reference images for each object by rendering the 3D model with different 3D orientation. We randomly sample $\mathbf{R}_{ref}$ using a 6D continuous representation~\cite{zhou2019continuity}, which results in better coverage of the orientation space. Following~\cite{wohlhart2015learning,park2020latentfusion,sundermeyer2020multi}, we crop the objects from the source images using the provided bounding boxes.

We compare our method with both a hand-crafted approach, i.e., HOG~\cite{dalal2005histograms}, and deep learning ones, i.e., LD~\cite{wohlhart2015learning}, NetVLAD~\cite{arandjelovic2016netvlad}, PFS~\cite{xiao2019pose}, MPE~\cite{sundermeyer2020multi}, and GDR-Net~\cite{wang2021gdr}. Note that DeepIM~\cite{li2018deepim} and LatentFusion~\cite{park2020latentfusion} are not evaluated since DeepIM requires an object pose initialization and LatentFusion needs additional depth information. We also exclude~\cite{sundermeyer2018implicit} because it requires training a separate autoencoder for every object, and thus cannot be used to estimate orientation for an unseen object without training a new autoencoder.

\subsection{Experiments on LineMOD and LineMOD-O}
We first conduct experiments on LineMOD~\cite{hinterstoisser2012model} and LineMOD-O~\cite{brachmann2014learning}. We split the cropped data into three non-overlapping groups, i.e., Split \#1, Split \#2, and Split \#3, according to the depicted objects. Deep learning models are trained on LineMOD and tested on both LineMOD and LineMOD-O. We augment training data by random occlusions when the evaluation is performed on LineMOD-O. In the case of unseen objects, we select one of the three groups as the testing set. We remove all images belonging to this group from training data, which ensures that no testing objects are observed in the training stage. In the case of seen objects on LineMOD, we separate $10\%$ from each group of this dataset for testing. We assume the object category to be unknown during testing, and therefore employ the evaluated methods to classify the object and then predict its 3D orientation.

We thus evaluate the tested methods in terms of both object classification accuracy and 3D orientation estimation accuracy. These are computed as 
\begin{equation}
\label{eq:eq9}
\text{Class. Acc.}=\left\{
\begin{array}{lcl}
1 &  \ \ \ \ & \text{if} \ \hat{C}_{ref}=C_{src} \\
0 & \ \ \ \ & \text{otherwise}
\end{array} \right.
\end{equation}
and 
\begin{equation}
\label{eq:eq10}
\text{Rota. Acc.}=\left\{
\begin{array}{lcl}
1 &  \ \ \ \ & \text{if} \ d(\hat{\mathbf{R}}_{ref},\mathbf{R}_{src}) < \lambda \ \text{and} \ \hat{C}_{ref}=C_{src} \\
0 & \ \ \ \ & \text{otherwise}
\end{array} \right.,
\end{equation}
respectively, with $\lambda=30^{\circ}$ a predefined threshold~\cite{xiao2019pose} and $d(\hat{\mathbf{R}}_{ref},\mathbf{R}_{src})$ the geodesic distance~\cite{wohlhart2015learning} between two rotation matrices. This distance is defined as 
\begin{align}
\label{eq:eq11}
d(\hat{\mathbf{R}}_{ref},\mathbf{R}_{src})=\text{arccos}(\frac{\text{tr}(\mathbf{R}_{src}^{\text{T}}\hat{\mathbf{R}}_{ref})-1}{2})/\pi.
\end{align}

We provide the results for LineMOD and LineMOD-O in Table~\ref{tab:linemod} and Table~\ref{tab:linemod_o}, respectively. As PFS assumes the object category is known, we only report its 3D orientation estimation accuracy. We replace the detection module in GDR-Net with the ground-truth bounding boxes in the presence of unseen objects since this detector cannot be used to detect unseen objects. Therefore, the classification accuracy of GDR-Net is not reported in this case. Being a traditional method, HOG does not differentiate seen and unseen objects, and thus achieves comparable results in both cases. However, its limited accuracy indicates that HOG suffers from other challenges, such as the appearance difference between real and synthetic images, and the presence of background and of occlusions. The previous retrieval-based methods, i.e, LD, NetVLAD, and MPE, achieve remarkable performance in the case of seen objects, but their accuracy significantly drops in the presence of unseen ones. This evidences that the global descriptors utilized in these approaches are capable of encoding 3D orientation information, but the described patterns are object specific, thus limiting the generalization ability of these methods to unseen objects. The performance of both PFS and GDR-Net also drops dramatically in the presence of unseen objects because the features extracted from 2D observations or 3D shapes remain strongly object dependent. Our method outperforms the competitors by a considerably large margin in the case of unseen objects, which demonstrates that the proposed patch-level image comparison framework generalizes better to unseen objects than previous works. This is because our use of patch-level similarities makes the network focus on local geometric attributes instead of high-level semantic information.

\begin{table}[!t]
	\begin{center}
	\scriptsize
	    \caption{\textbf{Experimental results on LineMOD~\cite{hinterstoisser2012model}.}}
	    \label{tab:linemod}
		\begin{tabular}{m{0.4cm}m{1.7cm}|m{1.12cm}<{\centering}m{1.12cm}<{\centering}|m{1.12cm}<{\centering}m{1.12cm}<{\centering}|m{1.12cm}<{\centering}m{1.12cm}<{\centering}|m{1.12cm}<{\centering}m{1.12cm}<{\centering}}
		    \hline
		    & & \multicolumn{2}{c|}{\textbf{Split \#1}} & \multicolumn{2}{c|}{\textbf{Split \#2}} & \multicolumn{2}{c|}{\textbf{Split \#3}} & \multicolumn{2}{c}{\textbf{Mean}} \\
		    & & Seen & \textbf{Unseen} & Seen & \textbf{Unseen} & Seen & \textbf{Unseen} & Seen & \textbf{Unseen} \\
		    
			\hline
            \multirow{6}*[-0.5pt]{\rotatebox{90}{Class. Acc. (\%)}} & HOG~\cite{dalal2005histograms} & 39.57 & 41.26 & 30.24 & 32.65 & 36.19 & 35.19 & 35.33 & 36.37 \\
            ~ & LD~\cite{wohlhart2015learning} & 99.17 & 52.53 & 99.02 & 47.85 & 97.94 & 30.60 & 98.71 & 43.66 \\
            ~ & NetVLAD~\cite{arandjelovic2016netvlad} & \textbf{100.00} & 68.36 & \textbf{100.00} & 52.30 & \textbf{100.00} & 47.22 & \textbf{100.00} & 55.96 \\
            ~ & PFS~\cite{xiao2019pose} & - & - & - & - & - & - & - & - \\
            ~ & MPE~\cite{sundermeyer2020multi} & 98.75 & 57.25 & 83.29 & 69.98 & 97.73 & 87.57 & 93.26 & 71.60 \\
            ~ & GDR-Net~\cite{wang2021gdr} & \textbf{100.00} & - & \textbf{100.00} & - & \textbf{100.00} & - & \textbf{100.00} & - \\
            ~ & Ours & \textbf{100.00} & \textbf{97.90} & 99.44 & \textbf{92.47} & 98.03 & \textbf{88.93} & 99.16 & \textbf{93.10} \\
			\hline
			\multirow{6}*[-1.5pt]{\rotatebox{90}{Rota. Acc. (\%)}} & HOG~\cite{dalal2005histograms} & 38.89 & 40.17 & 28.21 & 30.74 & 31.02 & 28.48 & 32.71 & 33.13 \\
             ~ & LD~\cite{wohlhart2015learning} & 94.50 & 8.63 & 89.57 & 12.47 & 91.47 & 5.22 & 91.85 & 8.77 \\
             ~ & NetVLAD~\cite{arandjelovic2016netvlad} & \textbf{100.00} & 36.11 & 98.66 & 20.33 & 99.35 & 23.38 & 99.34 & 26.61 \\
             ~ & PFS~\cite{xiao2019pose} & \textbf{100.00} & 6.31 & 99.19 & 6.65 & \textbf{99.46} & 5.54 & \textbf{99.55} & 6.17 \\
            ~ & MPE~\cite{sundermeyer2020multi} & 91.94 & 38.96 & 66.47 & 41.46 & 87.72 & 61.62 & 82.04 & 47.35 \\
            ~ & GDR-Net~\cite{wang2021gdr} & 99.89 & 4.61 & \textbf{99.28} & 4.82 & 99.31 & 5.02 & 99.49 & 4.82 \\
            ~ & Ours & 97.49 & \textbf{89.55} & 94.90 & \textbf{79.04} & 93.67 & \textbf{75.96} & 95.35 & \textbf{81.52} \\
            \hline
			\end{tabular}
	\end{center}
	\vspace{-15pt}
\end{table}
\begin{table}[!t]
	\begin{center}
	\scriptsize
	\caption{\textbf{Experimental results on LineMOD-O~\cite{brachmann2014learning}.}}
	\label{tab:linemod_o}
		\begin{tabular}{m{0.4cm}m{1.7cm}|m{1.12cm}<{\centering}m{1.12cm}<{\centering}|m{1.12cm}<{\centering}m{1.12cm}<{\centering}|m{1.12cm}<{\centering}m{1.12cm}<{\centering}|m{1.12cm}<{\centering}m{1.12cm}<{\centering}}
		    \hline
		    & & \multicolumn{2}{c|}{\textbf{Split \#1}} & \multicolumn{2}{c|}{\textbf{Split \#2}} & \multicolumn{2}{c|}{\textbf{Split \#3}} & \multicolumn{2}{c}{\textbf{Mean}} \\
		    & & Seen & \textbf{Unseen} & Seen & \textbf{Unseen} & Seen & \textbf{Unseen} & Seen & \textbf{Unseen} \\
			\hline
            \multirow{6}*[-0.5pt]{\rotatebox{90}{Class. Acc. (\%)}} & HOG~\cite{dalal2005histograms} & 0.60 & 0.60 & 0.23 & 0.23 & 36.20 & 36.20 & 12.34 & 12.34 \\
            ~ & LD~\cite{wohlhart2015learning} & 80.61 & 65.72 & \textbf{84.56} & 58.45 & 66.94 & 46.34 & 77.37 & 56.84 \\
            ~ & NetVLAD~\cite{arandjelovic2016netvlad} & \textbf{85.27} & 56.46 & 74.37 & 47.73 & \textbf{90.33} & 66.32 & 83.32 & 56.84 \\
            ~ & PFS~\cite{xiao2019pose} & - & - & - & - & - & - & - & - \\
            ~ & MPE~\cite{sundermeyer2020multi} & 83.29 & 56.07 & 55.26 & 45.08 & 70.72 & 57.55 & 69.76 & 52.90 \\
             ~ & GDR-Net~\cite{wang2021gdr}  & 84.48 & - & 83.81 & - & 89.19 & - & \textbf{85.83} & - \\
            ~ & Ours & 83.22 & \textbf{83.99} & 78.69 & \textbf{74.42} & 82.44 & \textbf{75.06} & 81.45 & \textbf{77.82} \\
			\hline
			\multirow{6}*[-1.5pt]{\rotatebox{90}{Rota. Acc. (\%)}} & HOG~\cite{dalal2005histograms} & 0.60 & 0.60 & 0.18 & 0.18 & 5.25 & 5.25 & 2.01 & 2.01 \\
            ~ & LD~\cite{wohlhart2015learning} & 32.21 & 6.25 & 26.56 & 3.26 & 24.57 & 4.57 & 27.78 & 4.69 \\
            ~ & NetVLAD~\cite{arandjelovic2016netvlad} & 51.60 & 24.32 & 42.20 & 18.05 & 36.56 & 18.84 & 43.45 & 20.40 \\
            ~ & PFS~\cite{xiao2019pose} & \textbf{71.40} & 6.25 & \textbf{60.88} & 13.15 & \textbf{54.67} & 4.68 & \textbf{62.32} & 8.73 \\
            ~ & MPE~\cite{sundermeyer2020multi}  & 40.47 & 22.56 & 27.31 & 5.20 & 35.06 & 18.22 & 34.28 & 15.33 \\
            ~ & GDR-Net~\cite{wang2021gdr} & 63.37 & 3.12 & 55.31 & 2.97 & 49.91 & 2.39 & 56.20 & 2.83 \\
            ~ & Ours & 64.92 & \textbf{60.75} & 56.51 & \textbf{52.41} & 52.47 & \textbf{37.85} & 57.97 & \textbf{50.34} \\
            \hline
			\end{tabular}
	 \end{center}
	 \vspace{-20pt}
\end{table}
\subsection{Experiments on T-LESS}
To further evaluate the generalization ability to unseen objects, we conduct an experiment on T-LESS~\cite{hodan2017t}. In this case, all deep learning approaches, including ours, were trained on LineMOD, and tested on the Primesense test scenes of T-LESS. As the objects' appearance and shape in T-LESS are significantly different from the ones in LineMOD, this experiment provides a challenging benchmark to evaluate generalization. As in~\cite{hodavn2016evaluation,sundermeyer2018implicit,sundermeyer2020multi}, we use $err_{vsd} \le 0.3$ as a metric on this dataset. Note that we do not use the refinement module in~\cite{sundermeyer2018implicit,sundermeyer2020multi} since it could be applied to all evaluated methods and thus is orthogonal to our contributions. As we concentrate on 3D object orientation estimation, we only consider the error of rotation matrices when computing $err_{vsd}$.  

The results of all the methods are provided in Table~\ref{tab:tless}. For this dataset, the 3D orientation estimates of all the previous deep learning methods are less accurate than those of the traditional approach, i.e., HOG. This shows that the models pretrained on LineMOD are unreliable when tested on T-LESS. By contrast, our method outperforms both hand-crafted and deep learning competitors. It evidences that our method can still effectively estimate 3D orientation for unseen objects even when the object's appearance and shape entirely differ from those in the training data.
\begin{table}[!t]
    \begin{center}
    \scriptsize
    \caption{\textbf{Experimental results on T-LESS~\cite{hodan2017t}.} All deep learning methods were trained on LineMOD and tested on the Primesense test scenes of T-LESS. We use $err_{vsd} \le 0.3$ as a metric.}
    \label{tab:tless}
        \begin{tabular}{c|c|c|c|c|c|c|c}
            \hline
            Method & HOG~\cite{dalal2005histograms} & LD~\cite{wohlhart2015learning} & NetVLAD~\cite{arandjelovic2016netvlad} & PFS~\cite{xiao2019pose} & MPE~\cite{sundermeyer2020multi} & GDR-Net~\cite{wang2021gdr} & Ours \\
            \hline
            Rota. Acc. (\%) & 74.22 & 24.19 & 56.46 & 17.92 & 66.88 & 11.89 & \textbf{78.73} \\
            \hline
        \end{tabular}
    \vspace{-10pt}
    \end{center}
\end{table}
\subsection{Ablation Studies}
\label{sec:abl}
\begin{figure}[!t]
	\centering
	\subfigure[]
	{\includegraphics[height=3cm]{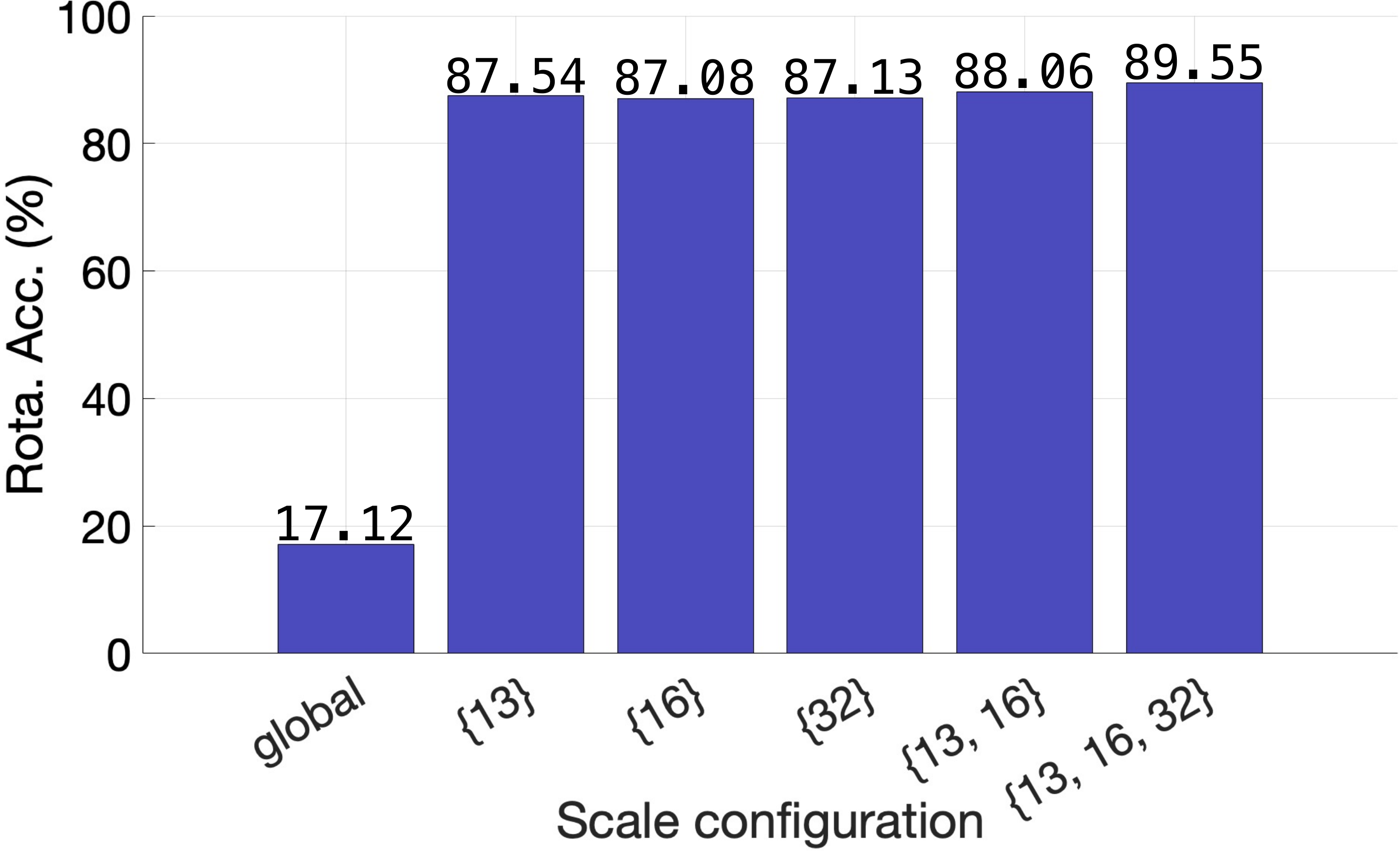}\label{fig:scale_a}} 
	\hspace{4em}
	\subfigure[]
	{\includegraphics[height=3cm]{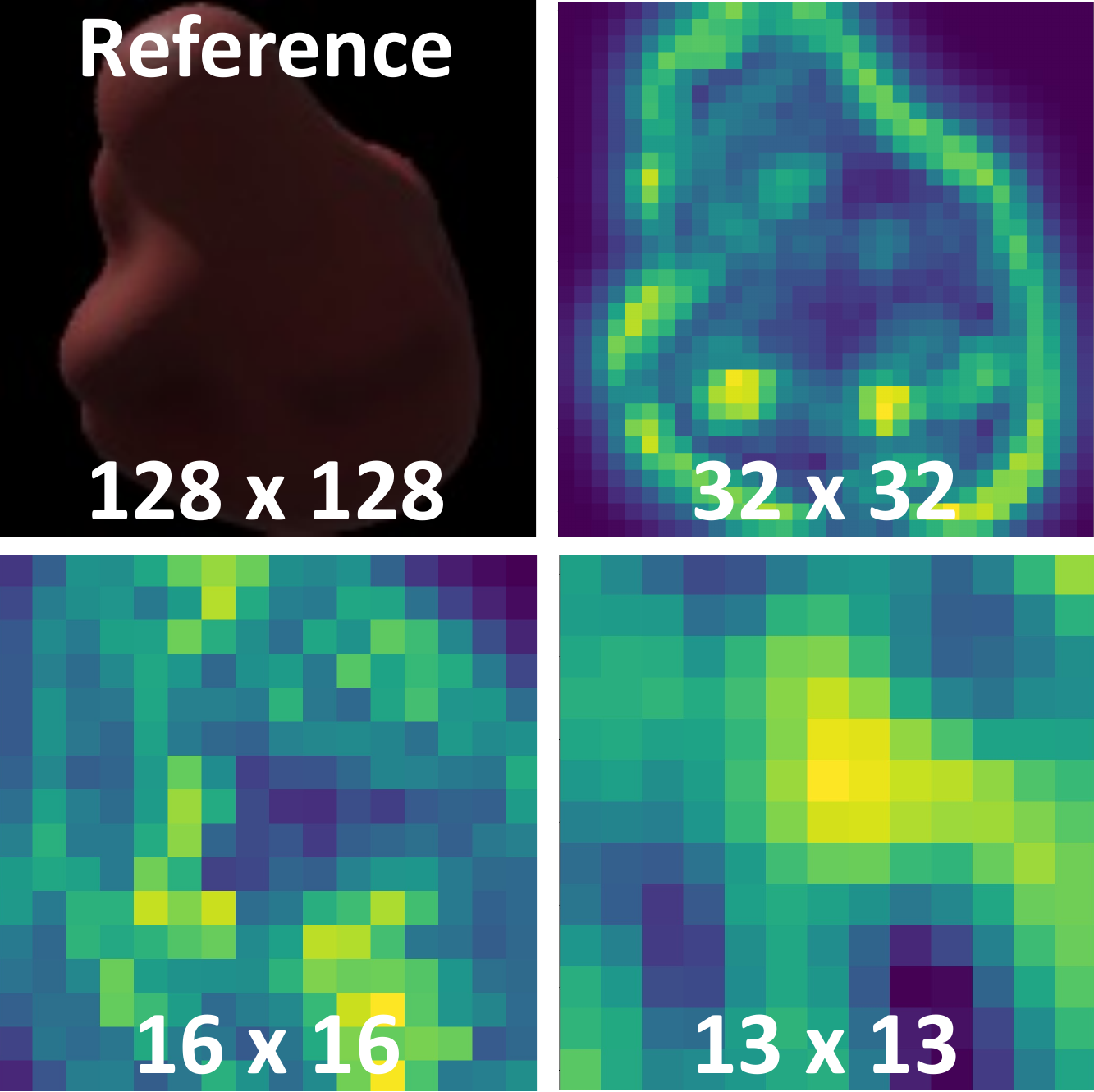}\label{fig:scale_b}}
	\caption{\textbf{Importance of local comparisons.} (a) Results of our method with different scale configurations on the unseen objects of LineMOD Split \#1. ``Global'' indicates a baseline that averages the smallest feature map into a global descriptor ($\mathbb{R}^{13\times13\times C} \rightarrow \mathbb{R}^{C}$) for retrieval while maintaining the other components, except for the adaptive fusion, unchanged. (b) Feature maps used for retrieval.}
	\label{fig:scale}
	 \vspace{-10pt}
\end{figure}

\noindent\textbf{Local Comparisons.} One of the key differences between our method and existing works~\cite{wohlhart2015learning,arandjelovic2016netvlad,sundermeyer2020multi} is our use of the local comparisons during retrieval. Fig.~\ref{fig:scale_a} demonstrates the importance of local features in our framework. We start from a ``global'' baseline in which we average the smallest feature map along the spatial dimensions to form a global descriptor ($\mathbb{R}^{13\times13\times C} \rightarrow \mathbb{R}^{C}$), which is used to compute the cosine similarity for retrieval. We keep the other components unchanged, except for the adaptive fusion. This baseline shows inferior generalization to unseen objects, while the performance significantly increases when local features are utilized. This observation indicates the importance of local comparisons for the unseen-object generalization. Moreover, the combination of multi-scale features also positively impacts the results, because it yields the ability to mix local geometric information at different scales, as illustrated by Fig.~\ref{fig:scale_b}.
\\

\begin{figure}[!t]
	\centering
	\subfigure[]
	{\includegraphics[height=2.2cm]{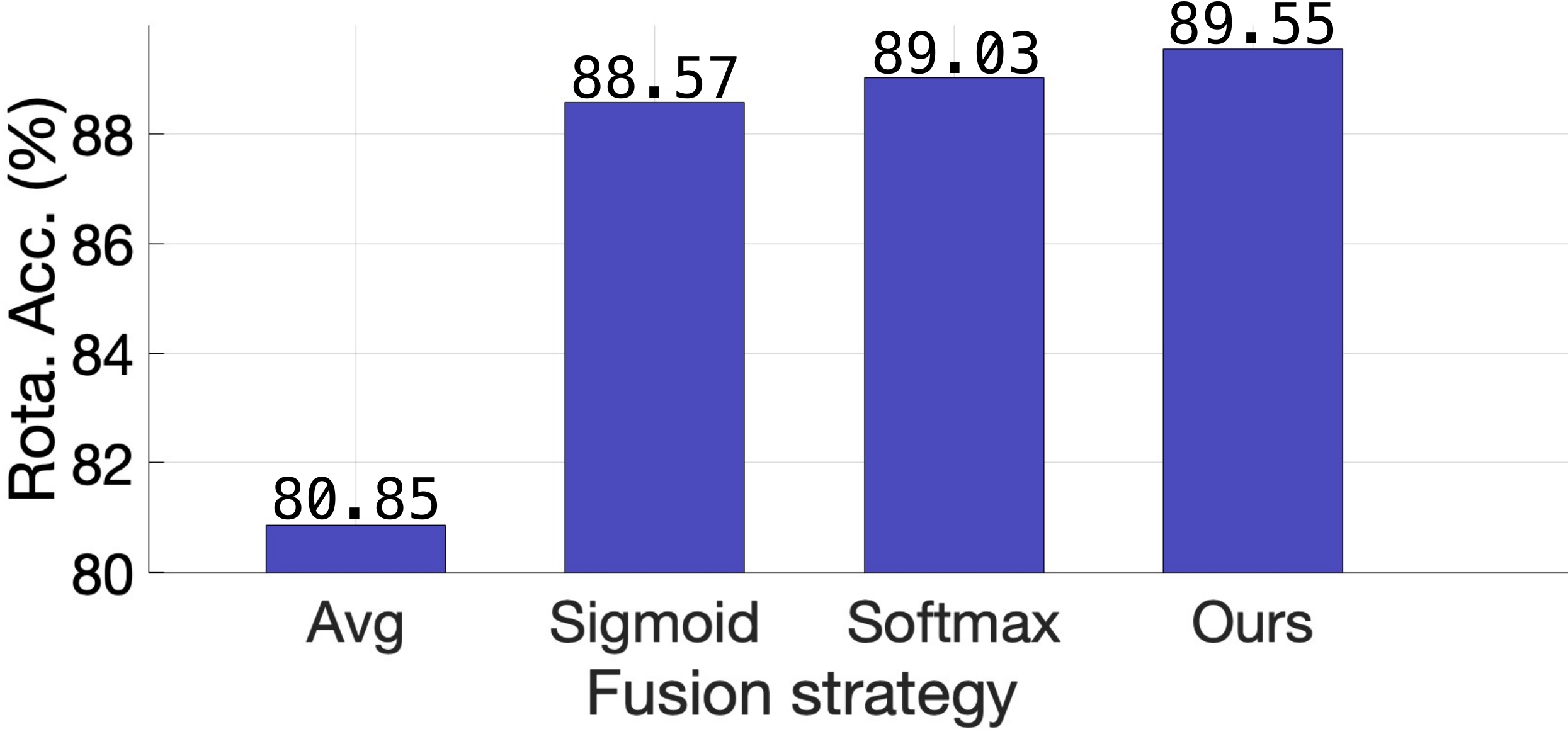}\label{tab:fusion}} 
	\subfigure[]
	{\includegraphics[height=2.2cm]{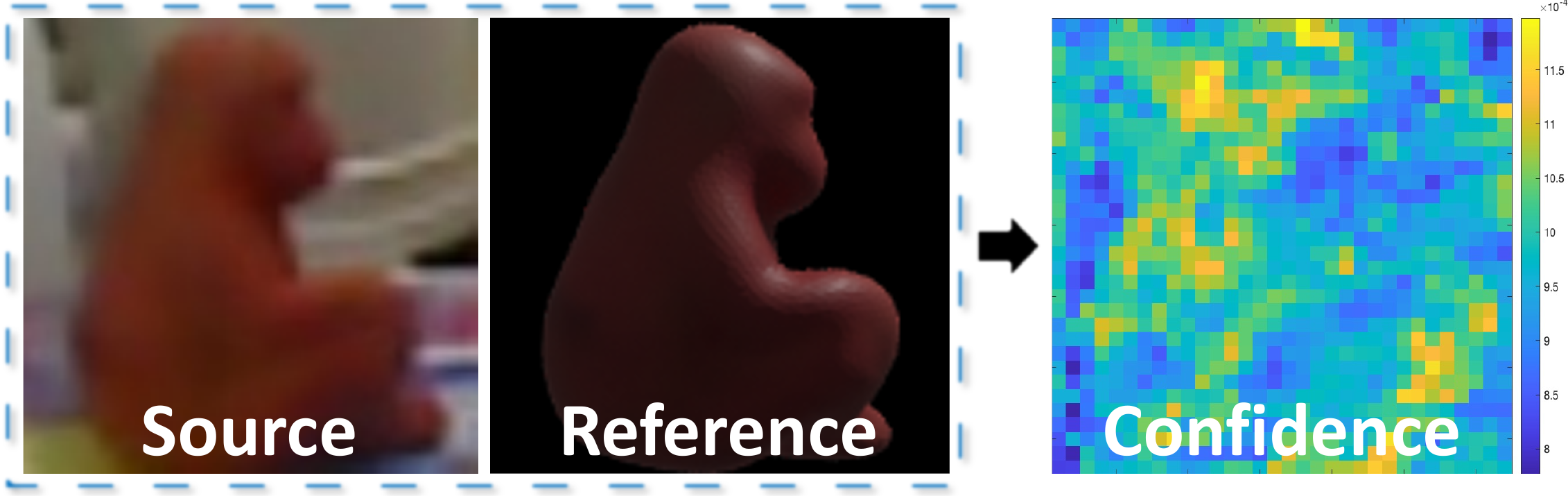}\label{fig:fusion}}
	\vspace{-10pt}
	\caption{(a) Our adaptive fusion significantly outperforms a simple averaging strategy, and yields a boost over the ``Sigmoid" and ``Softmax" alternatives. (b) Confidence map depicting the weights in Eq.~\ref{eq:eq6} obtained from the feature maps of the source and reference on the left. Yellow dots indicate local similarities with high confidence.}
	 \vspace{-5pt}
\end{figure}

\noindent\textbf{Adaptive Fusion.} To analyze the importance of the adaptive fusion module in our method, we introduce the following three baselines. ``Avg'' consists of replacing the adaptive fusion with a simple averaging process, estimating the image similarity score by averaging all per-element similarities over the pairs of feature maps. Furthermore, ``Sigmoid'' and ``Softmax'' involve using only the sigmoid or softmax function in Eq.~\ref{eq:eq6}, respectively. As shown in Fig.~\ref{tab:fusion}, our approach yields an $8.7\%$ increase in Rota. Acc. over ``Avg'', which demonstrates the superiority of our adaptive fusion strategy. The reason behind this performance improvement is that our module assigns different confidence weights to the local similarities, as illustrated in Fig.~\ref{fig:fusion}, which makes it possible to distinguish the useful information from the useless one. Moreover, the performance decreases (from 89.55\% to 88.57\% and 89.03\%) when separately employing sigmoid and softmax functions in Eq.~\ref{eq:eq6}, which indicates the effectiveness of combining local and global context.
\\

\noindent\textbf{Fast Retrieval.} We further conduct an experiment comparing the fast retrieval with a greedy search strategy. The greedy search compares the source image with every reference in the database during retrieval. To achieve a fair comparison, we divide all references into different groups with a group size of $1024$ and perform parallel estimation over the data in each group for the greedy search. As shown in Table~\ref{tab:fast}, although the greedy search achieves a better 3D orientation estimation accuracy, the time consumption (30.74s) is not affordable in practice. Note that LineMOD contains $13$ objects and we generate $10,000$ reference images for each object. Therefore, the image comparison is performed $130,000$ times for each source image in the greedy search, which results in an enormous time consumption. Note that we could not execute the comparisons w.r.t. all references in parallel because the NVIDIA Tesla V100 GPU could not store all the $130,000\times3$ feature maps. By contrast, our fast retrieval algorithm reduces the number of comparisons in two aspects: First, $\mathbf{I}_{src}$ is only compared with the initial anchors of each object for category recognition, reducing the complexity from $O(NR)$ to $O(Nk_{ac})$; second, the retrieval within the references that contain the recognized object for 3D orientation estimation only compares $\mathbf{I}_{src}$ with dynamically updated anchors, which reduces the complexity from $O(R)$ to $O(k_{ac})$.
\\
\begin{table}[!t]
    \begin{center}
    \caption{\textbf{Comparison between the fast retrieval and greedy search.}  We report the 3D orientation  accuracy and the test time on the unseen objects of LineMOD Split \#1.}
    \label{tab:fast}
        \begin{tabular}{c|c|c}
            \hline
            Method & Fast Retrieval & Greedy Search \\
            \hline
            Rota. Acc. (\%) & 89.55 & 95.93 \\
            Time (s) & 0.42 & 30.74 \\
            \hline
        \end{tabular}
    \end{center}
    \vspace{-18pt}
\end{table}

\begin{figure}[!t]
	\centering
	\includegraphics[width=0.5\linewidth]{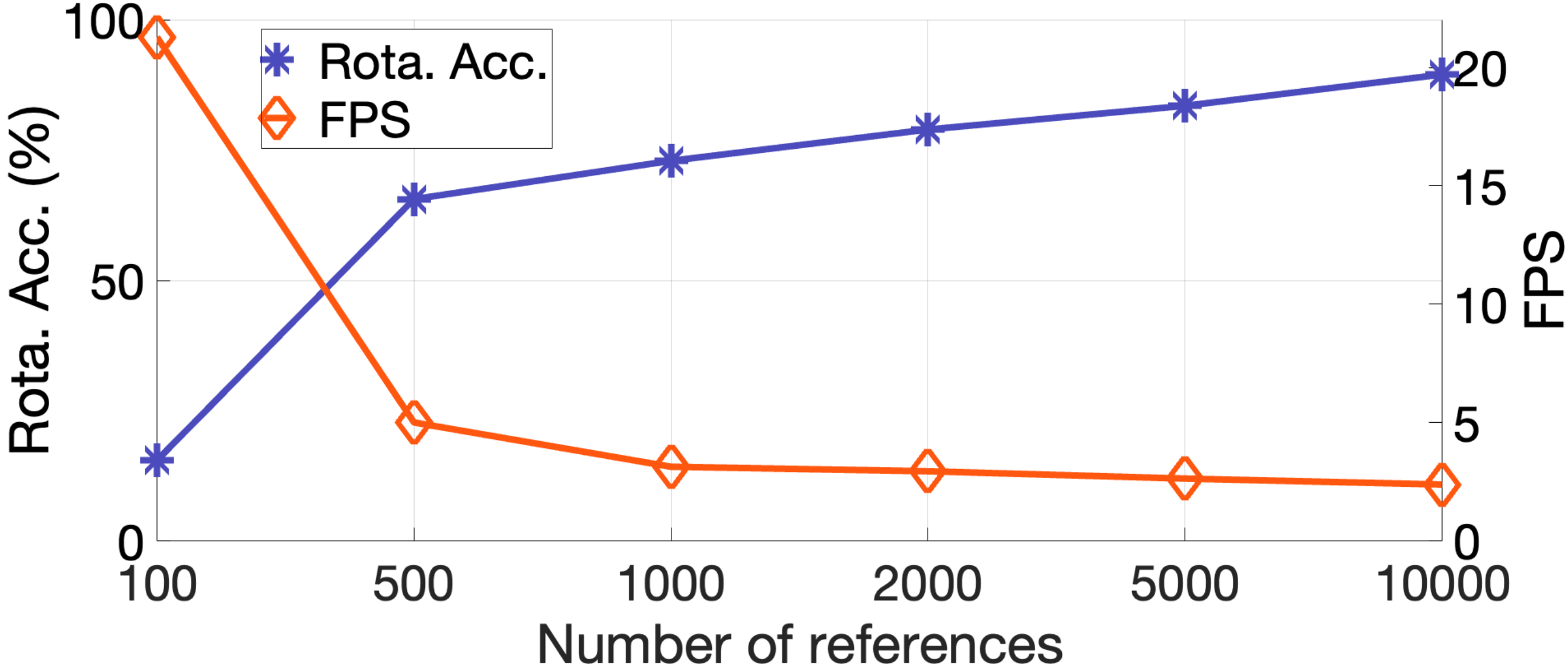}
	\caption{\textbf{Influence of the number of references.} We vary the number of references for each object and report the results on the unseen objects of LineMOD Split \#1.}
	\label{fig:ref_num}
\end{figure}

\noindent\textbf{Number of References.} Intuitively, we expect the number of reference images to be positively correlated with the 3D object orientation estimation accuracy while negatively correlated with the retrieval speed. To shed more light on the influence of the number of reference images, we evaluate our method with a varying number of references. As shown in Fig.~\ref{fig:ref_num}, as expected, as the number of reference images increases, Rota. Acc. increases while FPS decreases. Therefore, one can flexibly adjust the number of references in practice according to the desired accuracy and efficiency.
\\


\noindent\textbf{Effectiveness of the individual components.} Finally, we conduct an ablation study to further understand the importance of the other individual components in our method, i.e., local normalization and the weighted-infoNCE loss. As shown in Table~\ref{tab:ablation}, each of these two components has a positive impact on the 3D orientation estimation accuracy, and the optimal performance is achieved by leveraging both of them. 
\\

\begin{table}[!t]
    \begin{center}
    \caption{\textbf{Influence of the individual components.}}
    \label{tab:ablation}
        \begin{tabular}{c|c|c}
            \hline
            Local Norm. & Weighted infoNCE & Rota. Acc. (\%) \\
            \hline
            \xmark & \xmark & 85.91 \\
            \cmark & \xmark & 89.25 \\
            \xmark & \cmark & 87.14 \\
            \cmark & \cmark & \textbf{89.55} \\
            \hline
        \end{tabular}
    \end{center}
    \vspace{-20pt}
\end{table}

\vspace{-5pt}
\section{Conclusion}
In this paper, we have presented a retrieval-based 3D orientation estimation method for \emph{previously-unseen} objects. Instead of representing an image as a global descriptor, we convert it to multiple feature maps at different resolutions, whose elements represent local patches of different sizes in the original image. We perform retrieval based on patch-level similarities, which are adaptively fused into a single similarity score for a pair of images. We have also designed a fast retrieval algorithm to speed up our method. Our experiments have demonstrated that our method outperforms both traditional and previous learning-based methods by a large margin in terms of 3D orientation estimation accuracy for unseen objects. In future work, we plan to extend our method to full 6D pose estimation of unseen objects.
\\

\noindent\textbf{Acknowledgments.} This work was funded in part by the Swiss National Science Foundation and the Swiss Innovation Agency (Innosuisse) via the BRIDGE Discovery grant 40B2-0\_194729.
\clearpage

%
%
\bibliographystyle{splncs04}
\bibliography{egbib}
\end{document}

%% file: defs.tex
\newif\ifdraft
\draftfalse
\drafttrue

\ifdraft

 \newcommand{\MS}[1]{{\color{green}{\bf MS: #1}}}

 \newcommand{\YH}[1]{{\color{blue}{\bf YH: #1}}}

\else
 \newcommand{\PF}[1]{}
 
 \newcommand{\KY}[1]{}
 
 \newcommand{\MS}[1]{}
 
 \newcommand{\ZD}[1]{}
 
 \newcommand{\YH}[1]{}
 
 \newcommand{\SPE}[1]{}
 
 \newcommand{\WJ}[1]{}
 
\fi

